\documentclass[english]{cccconf}
\usepackage[comma,numbers,square,sort&compress]{natbib}
\usepackage{amsmath}
\usepackage{amssymb}
\begin{document}

\title{A sentiment analysis model for car review texts based on adversarial training and whole word mask BERT}
\author{Xingchen Liu\aref{scs}, 
        Yawen Li*\aref{sem},
        Yingxia Shao\aref{scs},
        Ang Li\aref{scs},
        Jian Liang\aref{didi}
        }


\affiliation[scs]{School of Computer Science (National Pilot School of Software Engineering), Beijing University of Posts and Telecommunications, Beijing Key Laboratory of Intelligent Telecommunication Software and Multimedia, P.~R.~China.}
\affiliation[sem]{School of Economics and Management, Beijing University of Posts and Telecommunications, P.~R.~China}
\affiliation[didi]{Didi Research Institute, Didi Chuxing, P.~R.~China.}
\maketitle

\begin{abstract}
In the field of car evaluation, more and more netizens choose to express their opinions on the Internet platform, and these comments will affect the decision-making of buyers and the trend of car word-of-mouth. As an important branch of natural language processing (NLP), sentiment analysis provides an effective research method for analyzing the sentiment types of massive car review texts. However, due to the lexical professionalism and large text noise of review texts in the automotive field, when a general sentiment analysis model is applied to car reviews, the accuracy of the model will be poor. To overcome these above challenges, we aim at the sentiment analysis task of car review texts. From the perspective of word vectors, pre-training is carried out by means of whole word mask of proprietary vocabulary in the automotive field, and then training data is carried out through the strategy of an adversarial training set. Based on this, we propose a car review text sentiment analysis model based on adversarial training and whole word mask BERT(ATWWM-BERT).
\end{abstract}


\footnotetext{*Corresponding author: Yawen Li (warmly0716@126.com).}

\section{Introduction}

At present, there are many car information websites in China that provide platforms for publishing car reviews, such as Dongchedi, Autohome, etc. A large number of car-buying users post their car-buying experience and product usage experience on these websites. Therefore, in the face of a large number of text review data containing evaluation content such as automotive product performance, quality, and brand services, how to make full use of text mining and other related technologies to efficiently mine the user emotional information contained in the review text, and find out the user's demand for products. Emotional tendencies, so as to assist consumers to make accurate purchasing decisions, help auto companies to improve products and services in a targeted manner, and enhance corporate competitiveness are important issues that need to be solved at present~\cite{c01,c03,Kou18recommendation,Kou16Social}. In this paper, we optimize the emotion recognition in the car review domain by increasing the whole word mask of the automotive domain vocabulary and adversarial training to the pre-trained Bidirectional Encoder Representation from Transformers (BERT) model.

For emotion recognition in the automotive field, the current main methods are the method based on emotion dictionary, the method based on machine learning, the method based on deep learning, etc.\cite{c05,c07,c08,c10,Li14LPV}. In today's Internet environment, a large number of new words are generated every day, and a lot of manpower is required to construct the sentiment dictionary, and the sentiment dictionary-based method strongly relies on the sentiment dictionary, so it has great limitations \cite{c05,Yang15Ontology}. Sentiment classification based on machine learning requires manual labeling of text features, and human subjective factors will affect the final result; secondly, machine learning needs to rely on a large amount of data, which is prone to ineffective work, and the model is not efficient, and it is difficult to adapt to today's fast In the era of development, such methods often fail to make full use of the contextual information of contextual texts when performing sentiment analysis, which will affect the accuracy. The sentiment analysis method based on deep learning has been developed rapidly in recent years \cite{c07,c34,Meng16Consensus}, and the accuracy has been greatly improved, but it cannot accurately classify professional vocabulary in specific fields.

In this paper, we consider the special form of Chinese and its importance in natural language processing, introduce proper nouns in the automotive field, and construct a BERT with whole word mask. At the same time, due to a large number of redundant punctuation marks in the training data, modal words, and other noises, so based on this, we use adversarial training to train the BERT covered by the whole word to extract more robust vectors and bring them into the classification network, so that the model can better adapt to the car reviews generated by users. Has good practical value.
The main contributions of this paper are as follows:

\begin{itemize}
  \item A BERT algorithm based on whole word mask of automobile domain vocabulary is proposed, By introducing proper nouns in the automobile domain, supervised learning is performed, which is more suitable for classifying the sentiment tendency of automobile reviews.

  \item A method for adversarial training of the above model is proposed, which improves the anti-interference ability of the model by adding disturbance factors to the word Embedding and the final fully connected layer.

  \item Through experimental comparison, the BERT model with whole word mask based on adversarial training proposed in this paper is superior to the most cutting-edge sentiment analysis model in both the accuracy and F1 of car evaluation sentiment analysis.
\end{itemize}

\section{Related work}
For emotion recognition in the automotive field, the current main methods are the method based on emotion dictionary, the method based on machine learning, the method based on deep learning, etc. \cite{c01,c03,c05,c07,c15}.

The method of a sentiment dictionary is to construct a sentiment dictionary artificially first. The sentiment value of sentiment words in the document is obtained by using the sentiment dictionary, and the overall sentiment tendency of the document is determined by weighted calculation. This method can define the sentiment of words, which is easy to analyze and understand by readers. If the content of the dictionary is rich enough, a better effect of sentiment analysis can be obtained \cite{c10,c23}. For example, Yanghui Rao proposed an efficient algorithm and three pruning strategies to automatically construct a word-level sentiment dictionary for social sentiment detection. and other related dictionaries to expand the sentiment dictionary. The sentiment value of a microblog text can be obtained through the calculation of the weight. Finally, Weibo texts about a topic can be classified as positive, negative, and neutral \cite{c24}. However, the method based on the sentiment dictionary relies too much on the sentiment dictionary and always has the limitation of the dictionary. In today's information age, new words appear every day, and the maintenance of the dictionary requires a great cost \cite{c27}; at the same time, this The method also does not consider the relationship between words, there is no context, and the sentiment dictionary is not easy to transfer across multiple datasets. Gradually, researchers began to study sentiment analysis based on machine learning.

The sentiment analysis method based on machine learning is to use the traditional machine learning algorithm to extract the features of the text, and use the algorithm to fit the training set \cite{Li17Distributed}. Compared with the method of using an emotional dictionary, machine learning does not rely on manual construction, which saves a lot of manpower. Through the database, the thesaurus can be updated in time. In machine learning, K-Nearest Neighbor algorithm(KNN)\cite{c12,c32}, Naïve Bayesian (NB) \cite{c13} and Support Vector Machine (SVM) \cite{c14} are commonly used learning algorithms. SVM and NB are better for the classification of text data [15]. Rodrigo Moraes proposed an SVM-based sentiment analysis method at the document level, which finally achieved 84.1\% results on the Movies dataset \cite{c25}. V Narayanan proposed Fast and accurate sentiment classification using an enhanced Naive Bayes model, demonstrating that the combination of word n-grams and mutual information feature selection can significantly improve accuracy \cite{c26}. The sentiment classification method based on machine learning has certain progress compared with the construction of a sentiment dictionary, but it still needs to manually mark the text features, and human subjective factors will affect the final result; secondly, machine learning needs to rely on a large amount of data, it is easy to Ineffective work is generated, and the execution speed will be very slow. If the efficiency of the model is not high, it is difficult to adapt to the current era of information explosion. Such methods often fail to make full use of the contextual information of the contextual text when performing sentiment analysis will have an impact \cite{Xu13Image}.

The introduction of deep learning has greatly promoted the development of sentiment classification. The deep learning method mainly uses embedding and multi-layer neural network \cite{,Shi19collaborative,Li17Distributed,Lin09Average,Li13Region} for forwarding propagation, and finally maps it into a set of vectors. The introduction of deep learning models has achieved good results in the field of sentiment analysis \cite{c17}. At present, the mainstream deep learning models for sentiment analysis include BiLSTM \cite{c18} proposed by S Hochreiter and Transformer \cite{c19} proposed by A Vaswani et al. Zhongdu has achieved the achievement of SOTA. Deep learning training models consume a lot of computing power, and the emergence of transfer learning and pre-training models has made deep learning popular \cite{c28}. A pre-trained model refers to a model that has been trained on a dataset. Researchers hope that the model that has spent a lot of time training can be retained and fine-tuned on their own dataset. The latest pre-training models for sentiment analysis include: Embeddings from Language Models (ELMo) \cite{c20}, BERT, and other deep learning-based sentiment analysis methods have developed rapidly in recent years, and the accuracy has been greatly improved, but for specific fields The professional vocabulary cannot be accurately classified.

Adversarial training was originally proposed by Generative adversarial nets, which is a method of defending against malicious attacks in the image field to improve the robustness of the model \cite{c21,Fang20GAN,Li22Scientific}; then Miyato applied adversarial training to text classification tasks in the field of natural language processing. as a regularization method to improve the classification performance of the model \cite{c22,c29,c30,c31}

\section{Method}
We propose an auto review text sentiment analysis model based on adversarial training and whole-word mask BERT(ATWWM-BERT), which is mainly composed of two parts, which are the whole-word mask BERT model based on the car review domain (Section 3.1) and the adversarial learning-based text classification strategy (Section 3.2) composition. Next, each of them will be described in detail.

\subsection{Whole-word mask BERT model based on car review domain}
In Chinese natural language processing, since the BERT model adopts the word segmentation method of word granularity in Chinese, a sentence is divided into an array composed of words. For example, the word "Bi Ya Di" will be split into three words "Bi", "Ya" and "Di" when inputting. During pre-training, these words segmented by the tokenizer are randomly replaced by [MASK]. Obviously, such a pre-training method makes BERT unable to learn the semantic information in Chinese text very well. In this paper, the Bidirectional Encoder Representation from Transformers-whole word mask (BERT-wwm) pre-training model, When some words in a phrase related to the car review domain are covered by [MASK] during training, other words that belong to the same phrase will also be [MASK]. After the BERT pre-training based on the whole-word mask of the car review field, the BERT pre-training model is more adaptable to the task of the car review field and can extract more representation information in the car review field, which can better solve the car review field. semantic ambiguity and sparse key features.
The specific whole-word mask pre-training process in the field of car reviews is as follows:

The MLM model is used to randomly block 15\% of the entries in the car review corpus sentences, and then the model predicts what the removed entries are. For the randomly occluded words, the pre-training method of whole word mask in the field of car reviews is used:
i). 80\% of the word vectors are replaced by [MASK] when inputting, if the word is part of the relevant nouns in the field of car reviews, other characters that belong to the same word will also be covered accordingly;
ii). The other 15\% of the word vectors are replaced by other word vectors. Similarly, if the word is part of the relevant nouns in the field of car reviews, other characters that belong to the same word will also be replaced accordingly;
iii), 5\% of the word vector input remains normal. The generated samples of whole word mask in the field of car reviews are shown in Tables \ref{tab1}.

\begin{table}[!htb]
  \centering
  \caption{Generating Examples of Whole Word Mask in the Domain of Car Reviews}
  \label{tab1}
  \begin{tabular}{p{1.9cm}|p{5.7cm}}

    \hhline
     Approach
    &
    Result
    \\ 
    \hline
    original text         
    &
    My Bi Ya Di drove 70,000 kilometers and consumed 8 fuel per 100 kilometers. The space and control are also good. Very satisfied with the purchase. 
    \\
    \hline
    
    original BERT generation case     
    &
    My Bi [MASK] Di drove 70,000 kilometers and consumed 8 fuel per 100 kilometers. The space and control are also good. Very [MASK] with the purchase. 
    \\
    \hline
    the whole word covers the field of car reviews      
    & 
    My [MASK][MASK][MASK] drove 70,000 kilometers and consumed 8 fuel per 100 kilometers. The space and control are also good. Very [MASK] with the purchase. 
    \\
    \hline
  \end{tabular}
\end{table}

We have achieved the pre-training process for the whole word mask of the car review field through the above method. From Table \ref{tab1}, we can find that this pre-training method can extract more information in the car review field and better solve the problems that appear in the car review. Problems with sparse proper nouns and key features. For the downstream task of sentiment analysis, LSTM and SoftMax layers are added to the output layer to output it. The model structure is shown in Figure \ref{figmodel}:

\begin{figure}[!htb]
  \centering
  \includegraphics[width=\hsize]{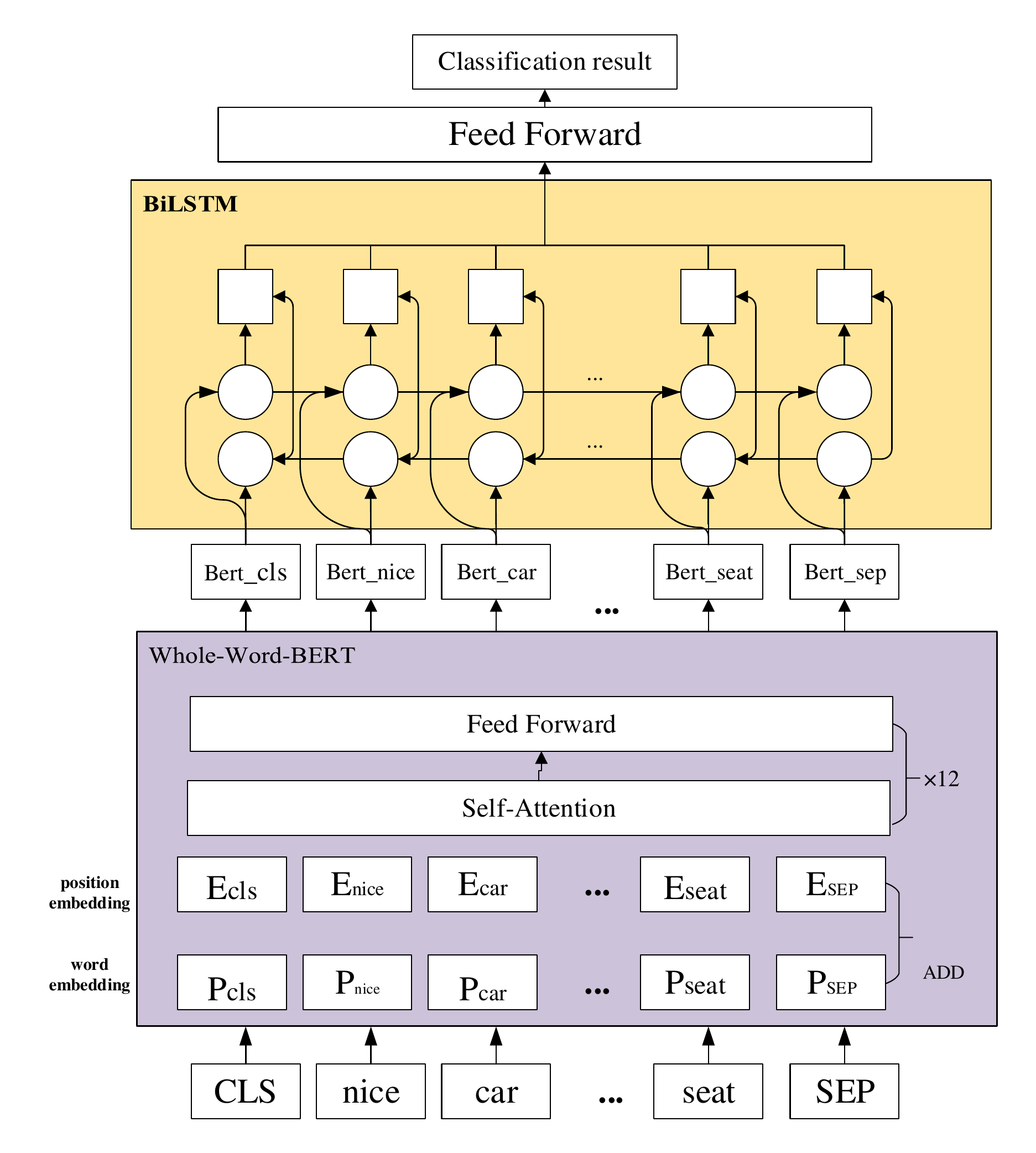}
  \caption{Model structure diagram}
  \label{figmodel}
\end{figure}

When the reviews and car review entities are input into the BERT pretrained model based on the whole word mask of the car review domain, the model processing is calculated as follows:

\begin{equation}
  \label{eq111}
  H{model} = BERT(Emb)
\end{equation}

\begin{equation}
  \label{eq112}
  L{model} = LSTM(H{model})
\end{equation}

\begin{equation}
  \label{eq113}
  M{model} = MLP(L{model})
\end{equation}

\begin{equation}
  \label{eq114}
  Label = SoftMax(M{model})
\end{equation}
Among them, MLP is a multi-layer perceptron network, which is used to compress the pre-trained features to the same feature dimension as the number of categories. The features are then classified by Softmax.
\subsection{Text Classification Strategy Based on Adversarial Training}
\begin{figure}[!htb]
  \centering
  \includegraphics[width=\hsize]{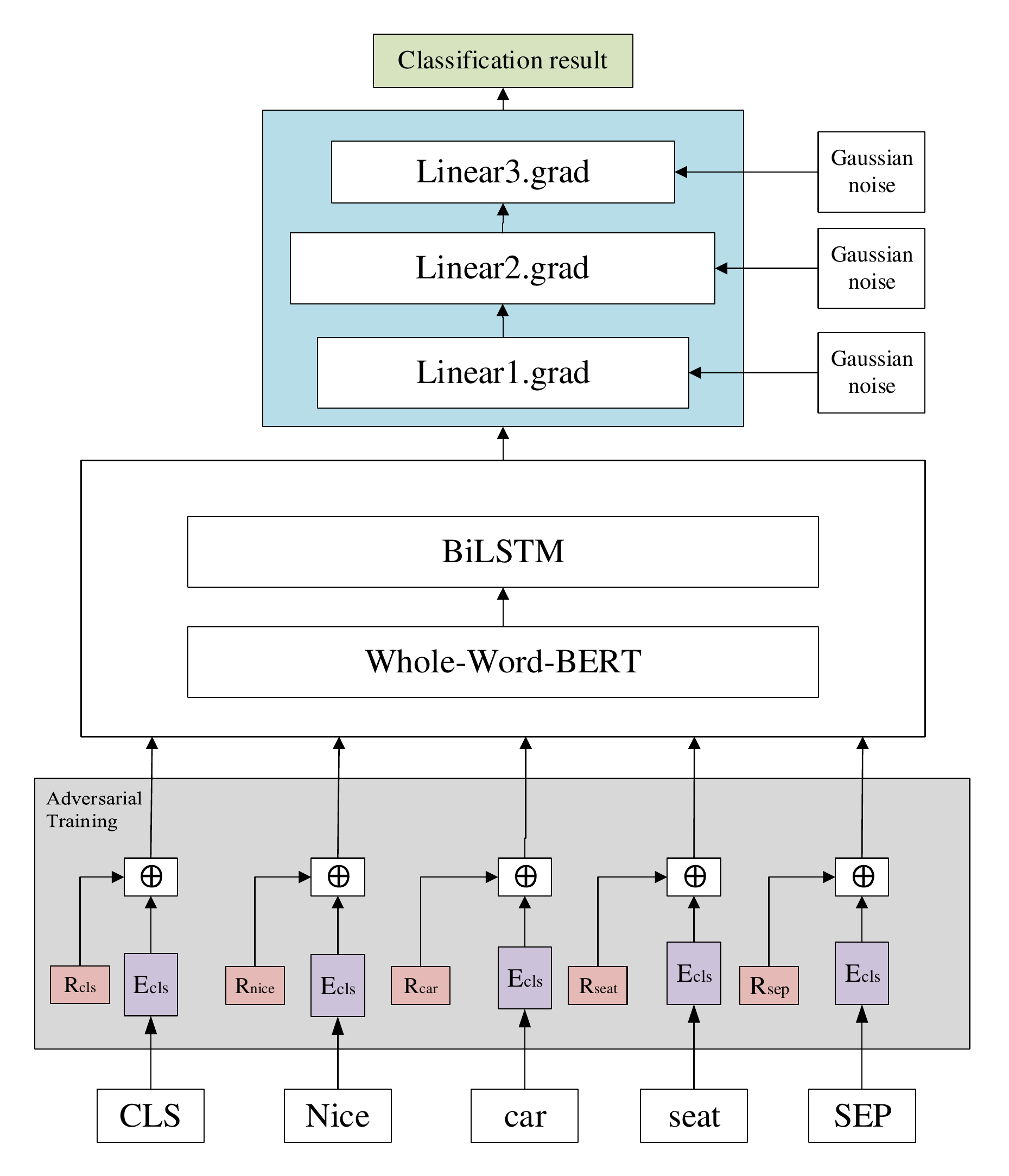}
  \caption{Adversarial training framework diagram}
  \label{figm}
\end{figure}

The adversarial training designed in the paper is mainly based on the addition of adversarial disturbances to the Embedding stage and the addition of the classification stage. It is mainly divided into two stages, called the representation learning stage and the text classification stage. The model architecture is shown in Figure \ref{figm}.

The goal of the representation learning phase is to use adversarial training methods to improve the quality of text representations, and to obtain better word vectors for the next text classification. In the representation learning model, forward and back propagation is used to train better words. By constructing adversarial examples, the generalization performance of the training model for correct classification can be improved, thereby improving the text accuracy.
We combine adversarial training methods and language models to improve the quality of text representations. In adversarial learning, we need to create the maximum noise, that is, the optimal adversarial disturbance, at which time the disturbance can maximize the loss of the model, namely:

\begin{equation}
\hat{r}=\underset{r,\|r\| \leqslant \varepsilon}{\arg \max }\left\{L_{\text {adv }}\left(\hat{x}_{1: \ell}, \theta^{\prime}, r\right)\right\}
\end{equation}

Among them, $ \varepsilon $ is used to constrain the size of the disturbance and is a hyperparameter. The optimal $ \varepsilon $ may be different for different tasks. We use the grid search strategy here, and finally, choose the value of $ \varepsilon $ to be 0.17.
in the above formula $ 
r=\left\{r_{i}\right\} \subset \mathrm{R}^{V} $ is the input value for the original text sequence $
x=\left\{x_{i}\right\} \subset \mathrm{R}^{V}$ interfere. That is, the value that finally enters the Bert model is:
\begin{equation}
\hat{x}_{1: \ell}=x_{1: \ell}+r_{1: \ell}
\end{equation}
The samples generated by adding perturbation to Embedding are called adversarial samples. $ L{adv}$ stands for adversarial loss, and its definition is similar to $L{lm} $. However, the value of this perturbation cannot be accurately calculated in the BERT network.
GoodFellow proposed an approximation algorithm,$L{adv} $ linearize around $ \hat{x}_{1: \ell}$. This method yields a non-iterative solution to $\hat{r}$

\begin{equation}
\hat{r}_{i}=\varepsilon \frac{g_{i}}{\left\|g_{2}\right\|_{2}}, g_{i}=\nabla_{x_{i}} L_{l m}\left(x_{1: \ell}, \theta^{\prime}\right)
\end{equation}
Among them, $\|g_{2}\|_{2} $presents L2 regularity.That is, the worst direction of the perturbation is the positive direction of the gradient of the loss in the input value. Therefore, the adversarial loss is defined as:
\begin{equation}
L_{\mathrm{adv}}\left(\hat{x}_{1: \ell}, \theta\right)=-\frac{1}{\ell} \sum_{i=1}^{\ell} \log p\left(\left(\hat{x}_{i} \mid \hat{x}_{1: i-1}\right) ; \theta^{\prime} ; \hat{r}\right)
\end{equation}
Adversarial training actually minimizes the worst-case error rate, i.e. minimizes the adversarial loss $ L_{{adv}}(x_{1: \ell}, \theta) $ So the final optimization objective of our model is:

\begin{equation}
\arg \min \left\{L_{l m}\left(x_{1: \ell}, \theta\right)+\lambda L_{\mathrm{adv}}\left(x_{1: \ell}, \theta\right)\right\}
\end{equation}

where $\lambda$ is a scalar hyperparameter that controls the balance of the two loss functions. The adversarial training method we use does not add additional computational overhead due to the non-iterative solution method and no back-propagation training.
The specific steps of the adversarial training method are shown as follows.

\begin{table}[!htb]
  \centering
  \label{tab44}
  \begin{tabular}{p{7.5cm}}
    \hhline
    Algorithm:Adversarial Training Algorithms           \\ \hline
    For bitch data set of x(t)         \\ 
    \quad1.Calculate the gradient $g=\bigtriangledown_{x}L_{lm}$\\
    \quad2.Adversarial perturbation $ \hat{r}_{i}=\varepsilon
    \frac{g_{i}}{\left\|g_{2}\right\|_{2}}$\\
    \quad3.Building the adversarial samples $\hat{x}_{1: \ell}=x_{1: \ell} + r_{1: \ell}$\\
    \quad4.Calculate adversarial losses$L_{adv}$\\
    \quad5.Minimizing loss function$L_{total}  = L_{lm}+\lambda L_{adv}$\\
    \quad6.Update parameters $\theta$ using gradient descent method\\
    \hline
  \end{tabular}
\end{table}

\section{Experiment}

\subsection{Datasets}
To demonstrate the effectiveness of our proposed model for sentiment analysis of car review texts based on adversarial training and whole word mask BERT. Since there is currently no Chinese sentiment analysis dataset related to the field of bus reviews, a sentiment analysis dataset in the field of car reviews is designed and constructed. Crawled comments from car review websites such as Autohome\footnote[1]{https://www.autohome.com.cn/} and Knowing Chedi\footnote[2]{https://www.dongchedi.com/}. 10543 car review news and comments were crawled through the Scrapy framework. After data cleaning and screening, 9947 items were selected as our dataset, and sentiment annotations were performed on them. We annotated a total of 3 categories of emotions: positive, neutral, and negative. The specific format is as follows: i) Positive emotion: It is recommended that 2.5T, CVT gearbox with dead weight, high body, China's road conditions are more complicated than Japan and the United States, the short-term power demand is still more, and the 2.5 fuel consumption is not high positive. ii) Neutral emotion: You can wait. The new Forester will be officially unveiled at the North American Auto Show at the end of the month. At that time, I will be considering whether to buy the new model or the old model. I am now waiting and watching. Footpads, chassis armor, chassis guards neutral. iii) Negative emotions: The Dodge I bought with my brother-in-law has high fuel consumption, almost power, and the interior space is not too large. For 7 seats, it costs 800 yuan for small maintenance. Negative. Among them, there are 3670 positive emotions, 3661 neutral emotions, and 2616 negative emotions. strip. We evenly distributed 9947 pieces of data according to sentiment, including 5000 pieces of training set, 2000 pieces of validation set, and 2947 pieces of test set.

\subsection{Evaluation Metrics}
This task uses the accuracy rate and Macro-F1 as evaluation indicators for evaluation, and the calculation formula for the accuracy rate is:

\begin{equation}
Acc = \frac{\sum_{i=1}^{m} TP_{i}}{Total}
\end{equation}

where $TP_{i}$ is the number of correct classifications for the i-th class. Total is the total number of data
Precision is the ratio of the number of correct predictions by the model to the total number of predictions of a certain class. It mainly measures the accuracy of the model.

\begin{equation}
Precision_{i} =\frac{TP_{i}}{TP_{i}+FP_{i}}
\end{equation}

Recall measures the rate at which a class is correctly classified by the model. It measures the recall rate of the model, where $TP_{i}$ is the number of correct predictions for a certain category, and $FN_{i} $ is the number of misidentified texts of this type. Its formula is:
\begin{equation}
Recall_{i} =\frac{T P_{i}}{T P_{i}+F N_{i}}
\end{equation}

So the formulas for average precision and average recall are as follows.

\begin{equation}
Precision_{mean}=\frac{\sum_{i=1}^{m}{ Precision }_{i}}{m}
\end{equation}

\begin{equation}
Recall_{mean}=\frac{\sum_{i=1}^{m}{ Recall }_{i}}{m}
\end{equation}

F1 is a comprehensive consideration of P and R. Its formula is.

\begin{equation}
F1_{Macro}=2 \frac{Recall_{m}*Precision_{m}}{Recall_{m}+Precision_{m}}
\end{equation}

\subsection{Settings}

The experimental parameters are set as follows: the maximum sequence length is set to 512, the maximum title and comment segment length is set to 510, the batch size is set to 16, the learning rate is set to 21e5, and the hidden layer size of the BERT model is set to 768, the optimization of the model The algorithm adopts Adam, which is a gradient descent algorithm with adaptive learning rate adjustment, which has the advantage of automatically adjusting the learning rate and accelerating the convergence speed. The dropout rate is set to 0.5, and the total number of epochs for training is 3 epochs. The perturbation factor e for adversarial training is set to 0.17.

\subsection{Compare models}
For a fine-tune mode and feature enhancement representation of our proposed BERT pre-training model, this section mainly compares three types of pre-training models, including basic BERT, BERT+LSTM and ERNIE:

\begin{itemize}
  \item BERT: The Transformer-based pre-training model published by Google has achieved SOTA on multiple NLP tasks. On this basis, we add a multi-layer perceptron to the model for emotion recognition.

  \item BERT+CNN: The output vector of bert is used as embedding inputs as the input of convolution; three different convolution kernels are used for convolution and pooling, and finally the three results are concat, and finally they are fully connected for classification.

  \item BERT+LSTM: Based on the BERT output vector, a LSTM model is attached to the head to obtain a time-based deeper vector for classification.
  \item ERNIE: It is a knowledge enhancement model proposed by Baidu in 2019. It learns real-world semantic relationships by modeling prior semantic knowledge such as entity concepts in a large amount of data.
\end{itemize}
To further demonstrate our model follow-up, we perform ablation experiments below.

\subsection{Experiment 1: Comparative experimental results}
We compared several models proposed in Section 4.4, in which the BERT model was used as the benchmark model, and the comparison results are shown in the Table \ref{tab223}.

\begin{table}[!htb]
  \centering
  \caption{Comparing the results of different models}
  \label{tab223}
  \begin{tabular}{p{3cm}|p{2.2cm}|p{2.2cm}}
    \hhline
    Model           & ACC & Macro-F1 \\ \hline
    BERT         & 68.56 & 64.89 \\ \hline
    BERT+LSTM       & 69.75 &65.76 \\ \hline
    BERT+CNN     & 68.98 & 65.12 \\ \hline
    ERNIE      & 70.23 & 67.45 \\ \hline
    \textbf{ATWWM-BERT}      & \textbf{77.68} & \textbf{74.75} \\ \hline
  \end{tabular}
\end{table}

Table \ref{tab223}. shows the comparison results of the model proposed in this chapter and other models.The bolded place is the best result. The final model achieved 74.75 and 77.76 in the macro-F1 and ACC evaluation indicators respectively, which is a significant improvement compared to the benchmark model BERT, and also has great advantages compared with the currently announced Chinese model (higher than ERNIE 7.3 points). It can be seen that our model can accurately analyze car reviews.
In order to more intuitively show the comparative effect of the algorithm in this paper and the cutting-edge sentiment classification algorithm. Table \ref{tab234} shows an example of the effect of whole word mask BERT based on adversarial learning on specific text classification. The bolded ones are the correctly classified results. The text of the classification is: Hong Qi is more official, although the fuel consumption is high, the driving is stable, and the interior is more atmospheric.

\begin{table}[!htb]
  \centering
  \caption{Comparing the results of different models}
  \label{tab234}
  \begin{tabular}{p{4cm}|p{3.5cm}}
    \hhline
    Model           & Classification result \\ \hline
    BERT         & neutral \\ \hline
    BERT+LSTM       & neutral \\ \hline
    BERT+CNN     & neutral \\ \hline
    ERNIE      & neutral \\ \hline
    \textbf{ATWWM-BERT}      & \textbf{positive} \\ \hline
  \end{tabular}
\end{table}

\begin{figure}[!htb]
  \centering
  \includegraphics[width=\hsize]{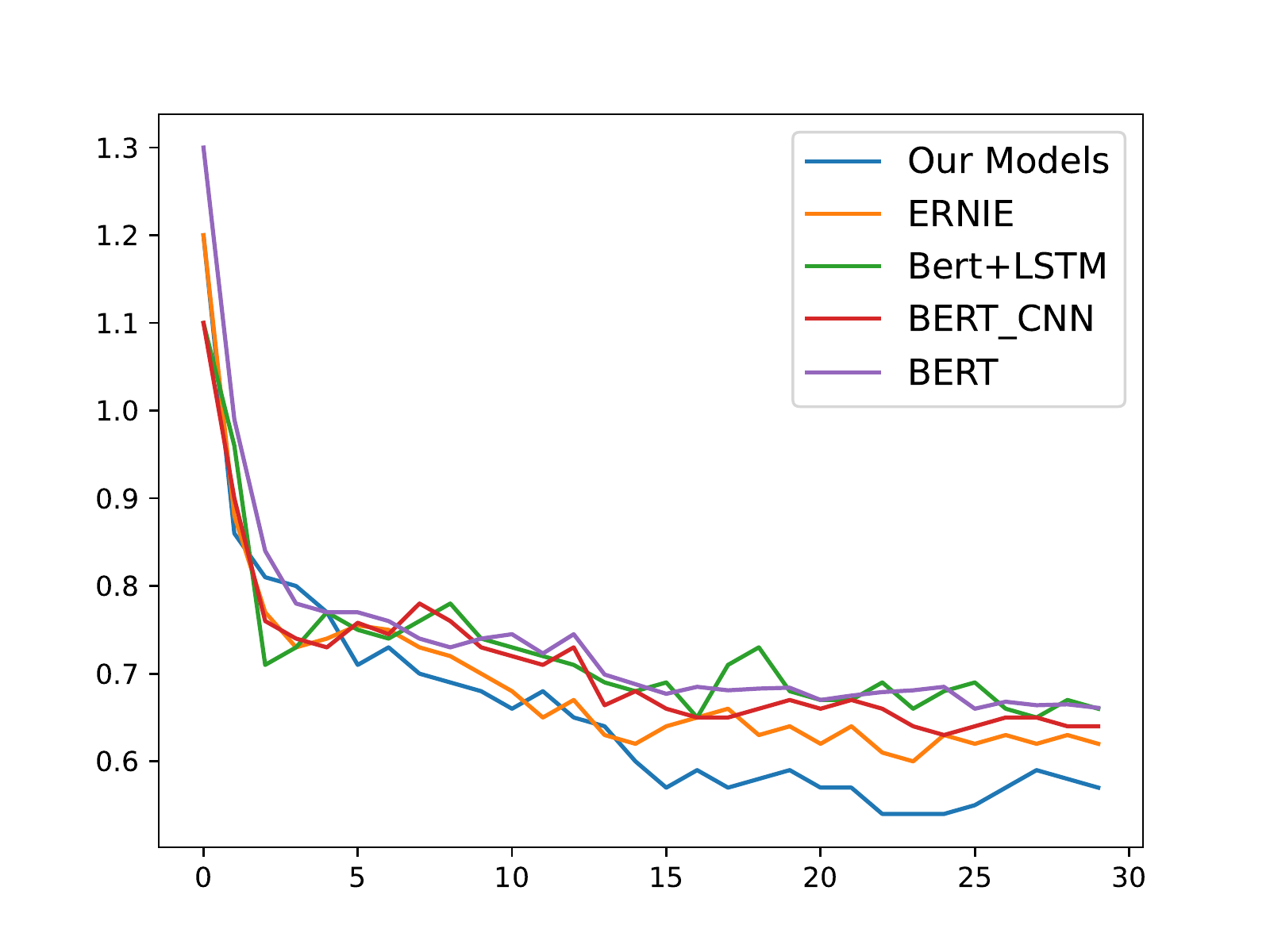}
  \caption{Loss drop graph of different models}
  \label{loss}
\end{figure}

Experiments show that the sentiment analysis model of car review text based on adversarial training and whole word mask BERT proposed in this chapter has better model expression, and has a better effect on the task of car review sentiment analysis. The Loss versus epoch curve is shown in Figure \ref{loss}.

It can be seen from the loss drop graph that the models tend to converge in the end. Among them, the BERT single model has the highest loss and the worst classification effect. The whole word mask BERT proposed by us based on adversarial learning has the best classification effect and the lowest loss. It is consistent with that shown in Table \ref{223}. It shows that our proposed model has good adaptability to the car review dataset.

\subsection{Experiment 2: Ablation experiment result}

In order to more clearly verify the collinearity of the two schemes proposed in the paper to the final model. In this paper, four ablation experimental schemes are mainly used to prove the experimental effect of the model, which are:
\begin{itemize}
  \item Whole-word mask BERT in the field of car reviews: The original BERT pre-training model was pre-trained with a whole word mask in the field of car reviews, and the position of the multi-layer perceptron was added to the model for sentiment classification.

  \item  BERT + adversarial training: The original BERT model is trained by adversarial training, and a fully connected layer is added at the end of the model for classification.

  \item  Whole-word mask BERT + adversarial training in the field of car review: The original BERT pre-training model was pre-trained against the whole word mask of the car review field, and the position of the multi-layer perceptron was added to the model for sentiment classification.
  \item  Whole-word mask in the field of car reviews BERT + adversarial training + LSTM: On the basis of the whole word mask and feature enhancement representation model in the field of car reviews, a bidirectional LSTM network is added to learn the sequence features between entries, and finally the whole word mask and feature enhancement representation model is added. Connection layer for classification.
\end{itemize}

\begin{table}[!htb]
  \centering
  \caption{Ablation experiment results}
  \label{tabab}
  \begin{tabular}{p{4cm}|p{1.7cm}|p{1.7cm}}
    \hhline
    Model           & ACC & Macro-F1 \\ \hline
    BERT         & 68.56 & 64.89 \\ \hline
    BERT-wwm      & 71.23 &69.32 \\ \hline
    BERT+ Adversarial     & 73.72 &69.32 \\ \hline
    BERT-wwm + Adversarial     & 75.72 & 73.78 \\ \hline
    \textbf{ATWWM-BERT}      & \textbf{77.68} & \textbf{74.75} \\
    \hline
  \end{tabular}
\end{table}

First, the experimental results of the BERT model show that it has the lowest value on both ACC and Macro-F1 results.The bolded place is the best result. 
At the same time, we tested the model after pre-training with whole word mask in the field of car reviews. Compared with the basic BERT language model, there was an improvement of 4.11 points in the results, confirming that the proposed pre-training of whole word mask in the field of car reviews was effective. effectiveness.
On the basis of the BERT model, we used the adversarial training method to train the data set for the BERT model. The results show that compared with the method without adversarial training, its ACC has improved by 2.76\%.
Based on the BERT model pre-trained with whole word mask in the field of car reviews, adversarial learning is added to enhance the expressive ability of the language model. Compared with the whole word mask BERT model in the field of car reviews, there is a 1.05\% improvement in value, which proves that adversarial learning is effective for sentiment analysis tasks in the field of car reviews. Finally, a bidirectional LSTM network is added to learn the sequence features between terms to form the final model of this chapter, which continues to improve the ACC value by 1.96\%. The validity of the sentiment analysis model proposed in this chapter based on whole-word mask and feature enhancement in the field of car reviews is verified.

\subsection{Error Analysis}
At the same time, this paper conducts an error analysis of the model, and we find that there is a lot of noise in car reviews, such as typos, redundant punctuation, and different people's understanding of different things (for example, some reviews car weight as an advantage, some as a disadvantage). When the length of the comment is greater than 512 characters, since the BERT model cannot extract all the features of the comment at one time, the model's sentiment prediction will be inaccurate due to missing features. For this reason, although the model in this paper is superior to other models, it still needs further improvement and further optimization in the model in sentiment analysis of long texts.

\section{Conclusion}
Aiming at the task of text sentiment analysis in the field of car reviews, in order to improve the classification accuracy in the professional field, this paper proposes a BERT preprocessing model based on whole word mask and adversarial training in the field of car reviews. In order to verify the advancedness and effectiveness of the proposed model, comparative experiments and ablation experiments were carried out on the model respectively. The experiments show that the model proposed in this paper has achieved higher classification accuracy in the task of sentiment analysis of car review texts. At the same time, the study also found that when the length of the comment is greater than 512 characters or the comment text has a lot of noise, because the BERT model cannot extract all the features of the comment at one time, the accuracy of the model recognition is limited. We speculate that if the method in this paper is applied to other domains, similar improvements in classification performance can be obtained.


\begin{thebibliography}{0}
\bibitem{c01}
He Yang. Research on Sentiment Classification and Evaluation of Car Reviews Based on Evidence Reasoning [D]. Hefei University of Technology, 2020. DOI: 10.27101/d.cnki.ghfgu.2020.001666.



\bibitem{c03}
Cheng Ran. Sentiment analysis of car review texts based on multi-label classification [D]. Donghua University, 2021. DOI: 10.27012/d.cnki.gdhuu.2021.000998.

\bibitem{Kou18recommendation}
Feifei Kou, Junping Du, Congxian Yang, Yansong Shi, Wanqiu Cui, Meiyu Liang, and Yue Geng. Hashtag recommendation based on multi-features of microblogs. Journal of Computer Science and Technology, 2018, 33(4): 711-726.

\bibitem{Kou16Social}
Feifei Kou, Junping Du, Yijiang He, Lingfei Ye. Social network search based on semantic analysis and learning. CAAI Transactions on Intelligence Technology, 1(4): 293-302, 2016. 



\bibitem{c05}
Wang Chundong, Zhang Hui, Mo Xiuliang, Yang Wenjun. A Review of Sentiment Analysis on Weibo [J]. Computer Engineering and Science, 2022, 44(01): 165-175.


\bibitem{c07}
He Yanxiang, Sun Songtao, Niu Feifei, Li Fei. A Deep Learning Model for Sentiment Enhancement of Weibo Sentiment Analysis [J]. Chinese Journal of Computers, 2017, 40(04): 773-790.

\bibitem{c08}
Devlin J, Chang M W, Lee K, et al. Bert: Pre-training of deep bidirectional transformers for language understanding[J]. arXiv preprint arXiv:1810.04805, 2018.


\bibitem{c10}
Wang Zhitao, Yu Zhiwen, Guo Bin, Lu Xinjiang. Sentiment Analysis of Chinese Weibo Based on Dictionary and Rule Set [J]. Computer Engineering and Applications, 2015, 51(08): 218-225.

\bibitem{Li14LPV}
Li M, Jia Y, and Du J. LPV control with decoupling performance of 4WS vehicles under velocity-varying motion[J]. IEEE Transactions on Control Systems Technology 2014, 22(5): 1708-1724.

\bibitem{Yang15Ontology}
Yuehua Yang, Junping Du, and Yuan Ping. Ontology-based intelligent information retrieval system[J]. Journal of Software, 26(7): 1675-1687, 2015.


\bibitem{Li17Distributed}
Li W, Jia Y, and Du J. Distributed extended Kalman filter with nonlinear consensus estimate[J]. Journal of the Franklin Institute, 2017, 354(17): 7983-7995.  

\bibitem{c12}
Peterson L E. K-nearest neighbor[J]. Scholarpedia, 2009, 4(2): 1883.

\bibitem{c32}
Sun B, Du J, Gao T. Study on the improvement of K-nearest-neighbor algorithm[C]. 2009 International Conference on Artificial Intelligence and Computational Intelligence, 4: 390-393, 2009.   

\bibitem{c34}
Hu W, Gao J, Li B, et al. Anomaly detection using local kernel density estimation and context-based regression[J]. IEEE Transactions on Knowledge and Data Engineering, 32(2): 218-233, 2018. 

\bibitem{Meng16Consensus}
Meng D, Jia Y, and Du J. Consensus seeking via iterative learning for multi‐agent systems with switching topologies and communication time‐delays[J]. International Journal of Robust and Nonlinear Control, 2016, 26(17): 3772-3790. 

\bibitem{c13}
Webb G I, Keogh E, Miikkulainen R. Naïve Bayes[J]. Encyclopedia of machine learning, 2010, 15: 713-714.

\bibitem{c14}
Ding Shifei, Qi Bingjuan, Tan Hongyan. A Review of Support Vector Machine Theory and Algorithm Research [J]. Journal of University of Electronic Science and Technology of China, 2011, 40(01): 2-10.

\bibitem{c15}
Li Ran, Lin Zheng, Lin Hailun, Wang Weiping, Meng Dan. A Review of Text Sentiment Analysis [J]. Computer Research and Development, 2018, 55(01): 30-52.


\bibitem{Xu13Image}
Xu L, Du J, Li Q. Image fusion based on nonsubsampled contourlet transform and saliency-motivated pulse coupled neural networks[J]. Mathematical Problems in Engineering, 2013.  

\bibitem{Shi19collaborative}
Shi C, Han X, Song L, et al. Deep collaborative filtering with multi-aspect information in heterogeneous networks[J]. IEEE transactions on knowledge and data engineering, 2019, 33(4): 1413-1425. 

\bibitem{Li17Distributed}
Li W, Jia Y, Du J. Distributed consensus extended Kalman filter: a variance-constrained approach[J]. IET Control Theory \& Applications, 11(3): 382-389, 2017.

\bibitem{Lin09Average}
Lin P, Jia Y, Du J, et al. Average consensus for networks of continuous-time agents with delayed information and jointly-connected topologies[C]. 2009 American Control Conference, 2009: 3884-3889.  

\bibitem{Li13Region}
Li Q, Du J, Song F, et al. Region-based multi-focus image fusion using the local spatial frequency[C]. 2013 25th Chinese control and decision conference (CCDC), 2013: 3792-3796.  

\bibitem{c17}
LeCun Y, Bengio Y, Hinton G. Deep learning[J]. nature, 2015, 521(7553): 436-444.
\bibitem{c18}
Xu G, Meng Y, Qiu X, et al. Sentiment analysis of comment texts based on BiLSTM[J]. Ieee Access, 2019, 7: 51522-51532.

\bibitem{c19}
Vaswani A, Shazeer N, Parmar N, et al. Attention is all you need[J]. Advances in neural information processing systems, 2017, 30.

\bibitem{c20}
Peters, M. E., Deep contextualized word representations, arXiv e-prints, 2018.
\bibitem{c21}
Goodfellow I, Pouget-Abadie J, Mirza M, et al. Generative adversarial nets[J]. Advances in neural information processing systems, 2014, 27.

\bibitem{Fang20GAN}
Fang Y, Deng W, Du J, Hu J. Identity-aware CycleGAN for face photo-sketch synthesis and recognition[J]. Pattern Recognition, 102: 107249, 2020.  

\bibitem{Li22Scientific}
Li A, Du J, Kou F, et al. Scientific and Technological Information Oriented Semantics-adversarial and Media-adversarial Cross-media Retrieval[J]. arXiv preprint arXiv:2203.08615, 2022.

\bibitem{c22}
Miyato T, Dai A M, Goodfellow I. Adversarial training methods for semi-supervised text classification[J]. arXiv preprint arXiv:1605.07725, 2016.
\bibitem{c23}
Rao Y, Lei J, Wenyin L, et al. Building emotional dictionary for sentiment analysis of online news[J]. World Wide Web, 2014, 17(4): 723-742.

\bibitem{c24}
Zhang S, Wei Z, Wang Y, et al. Sentiment analysis of Chinese micro-blog text based on extended sentiment dictionary[J]. Future Generation Computer Systems, 2018, 81: 395-403.

\bibitem{c25}
Moraes R, Valiati J F, Neto W P G O. Document-level sentiment classification: An empirical comparison between SVM and ANN[J]. Expert Systems with Applications, 2013, 40(2): 621-633.

\bibitem{c26}
Narayanan V, Arora I, Bhatia A. Fast and accurate sentiment classification using an enhanced Naive Bayes model[C]//International Conference on Intelligent Data Engineering and Automated Learning. Springer, Berlin, Heidelberg, 2013: 194-201.
\bibitem{c27}
Van Atteveldt W, van der Velden M A C G, Boukes M. The validity of sentiment analysis: Comparing manual annotation, crowd-coding, dictionary approaches, and machine learning algorithms[J]. Communication Methods and Measures, 2021, 15(2): 121-140.

\bibitem{c28}
Qiu X, Sun T, Xu Y, et al. Pre-trained models for natural language processing: A survey[J]. Science China Technological Sciences, 2020, 63(10): 1872-1897.
\bibitem{c29}
Liu P, Qiu X, Huang X. Adversarial multi-task learning for text classification[J]. arXiv preprint arXiv:1704.05742, 2017.
\bibitem{c30}
Garg S, Ramakrishnan G. Bae: Bert-based adversarial examples for text classification[J]. arXiv preprint arXiv:2004.01970, 2020.
\bibitem{c31}
Song L, Yu X, Peng H T, et al. Universal adversarial attacks with natural triggers for text classification[J]. arXiv preprint arXiv:2005.00174, 2020.


\end{thebibliography}
\end{document}